УДК 004.93

# АЛГОРИТМ РАСПОЗНАВАНИЯ КОНТУРА СОТОБЛОКА

[1]Кубриков Максим Викторович, канд. тех. наук, ведущий научный сотрудник;
[2]Сарамуд Михаил Владимирович, канд. тех. наук, старший научный сотрудник;
[3]Паулин Иван Алексеевич, инженер;
[4]Талай Евгений Петрович, инженер.

[1]ФГБОУ ВО «Сибирский государственный университет науки и технологий имени академика М.Ф. Решетнева», Красноярск, Россия, e-mail: gidroponika@mail.ru
[2]ФГБОУ ВО «Сибирский государственный университет науки и технологий имени академика М.Ф. Решетнева», Красноярск, Россия, e-mail: saramud@gmail.com
[3]ФГБОУ ВО «Сибирский государственный университет науки и технологий имени академика М.Ф. Решетнева», Красноярск, Россия, e-mail: ylanava13@mail.ru
[4]ФГБОУ ВО «Сибирский государственный университет науки и технологий имени академика М.Ф. Решетнева», Красноярск, Россия e-mail: il20596@mail.ru

*В статье рассмотрен алгоритм распознавания контура фрагментов сотоблока. Показана неприменимость готовых функций библиотеки OpenCV. Рассмотрены два предложенных алгоритма. Алгоритм прямого сканирования находит крайние белые пиксели на бинаризованном изображении, он адекватно работает на выпуклых формах изделий, однако не находит контур на вогнутых областях и в полостях изделий. Для решения этой проблемы предложен алгоритм сканирования с помощью скользящей матрицы, который корректно работает на изделиях любой формы.*

## Вступление

Для роботизированного комплекса раскроя сотоблочных изделий требуется определять контуры заготовки из которой будут вырезаться заготовки требуемой формы [1]. Так как сотоблок имеет неоднородную структуру (содержит множество контрастных граней внутри), стандартными функциями библиотеки OpenCV [2] решить данную задачу не представляется возможным. Из-за особой структуры, функция для поиска контуров findContours не может выделить единый контур сотоблока. На рисунке 1 представлены различные примеры применения данной функции к образцам сотоблоков.

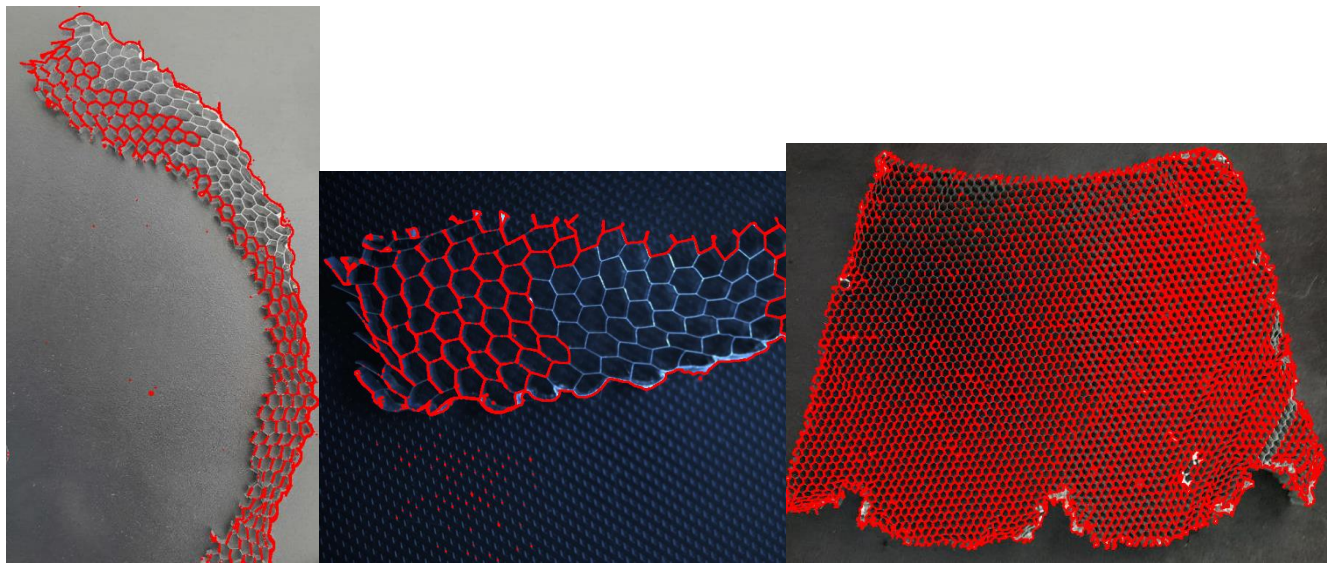

а) б) в)

*Рис. 1. Обнаружение контуров с помощью стандартных средств OpenCV*

Как можно видеть данная функция распознает соты как отдельные контуры. Для успешного нахождения внешнего контура сотоблока были разработаны 2 алгоритма. Оба алгоритма используют машинное зрение, но разный принцип обработки изображения.

### Алгоритм прямого сканирования.

Данный алгоритм производит "сканирование" бинаризованного изображения [3], которое производится по горизонтали, проверяем цвет каждого пикселя (белый или нет), когда встречаем белый пиксель считаем его за начало контура (левая сторона со стороны наблюдателя). Поскольку в сотоблоке идет чередование белых и черных пикселей для определение правой точки контура сотоблока необходимо считать количество черных пикселей, которые идут подряд. Если количество черных пикселей превышает определенное значение считаем последнюю белую точки правой границей сотоблока. На рисунке 2 показана схема работы данного алгоритма.

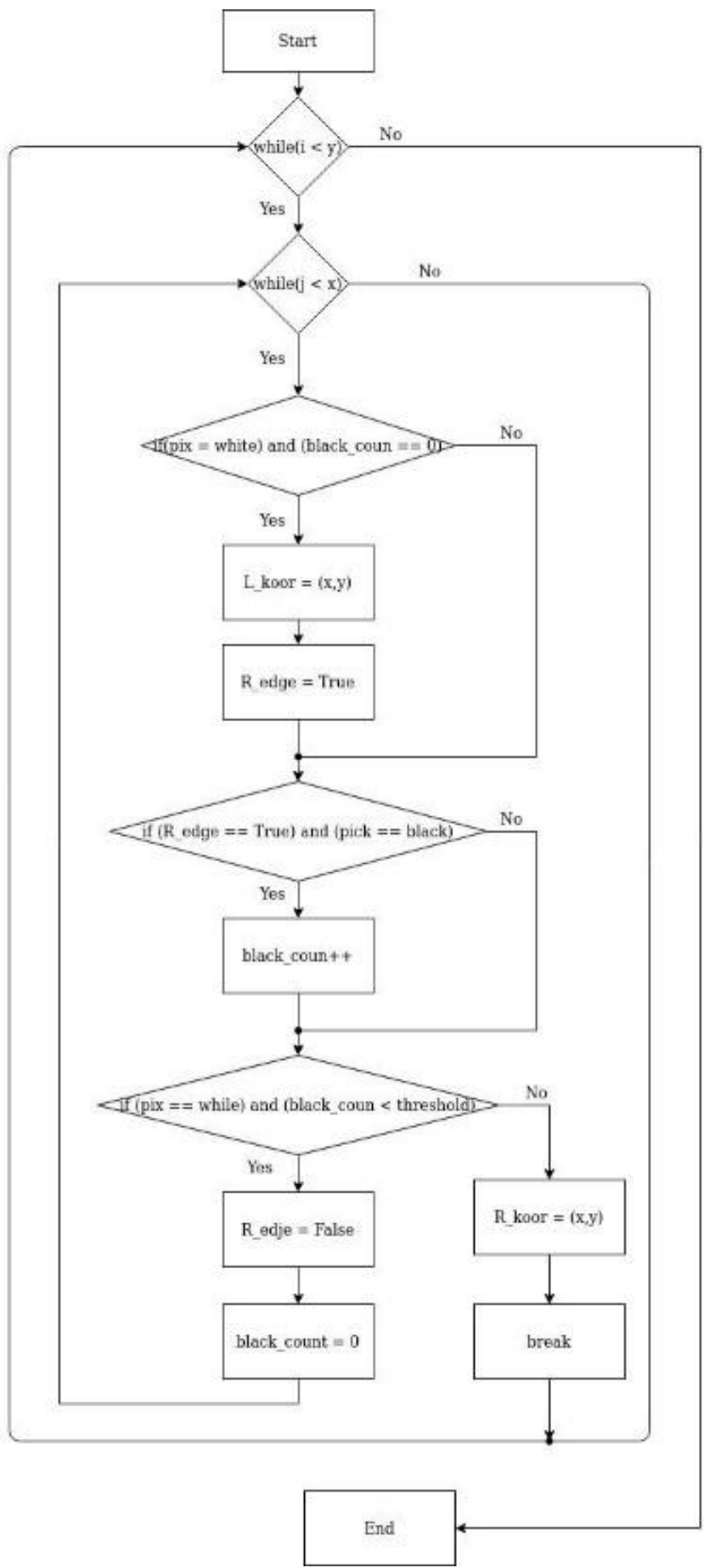

Рис. 2. Блок-схема алгоритма

На рисунке 3 показаны найденные точки, красным выделены точки левого края сотоблока, зеленым обозначены точки правого края.

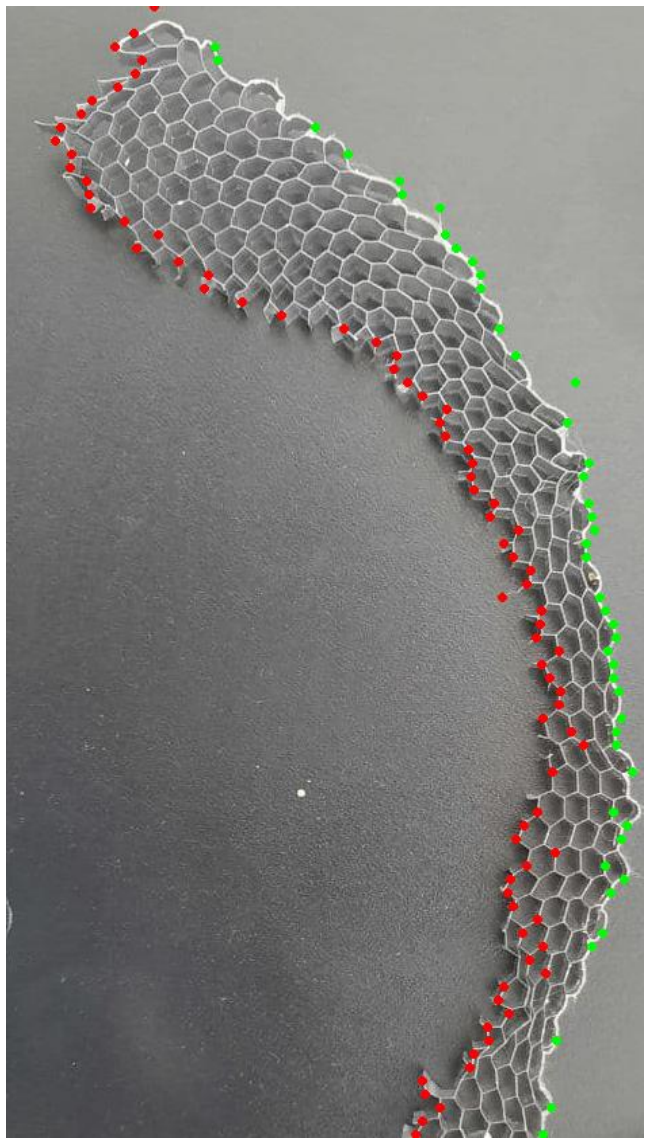

*Рис. 3. Точки контура сотоблока.*

Для того, чтобы не захватывать ложные точки в виде шумов мы проверяем расстояние между координатами оси X левой и правой точки. Если расстояние между ними меньше порога, то данные точки учитываться не будут. На рисунке 4 представлено бинаризованное изображение по которому идет сканирование.

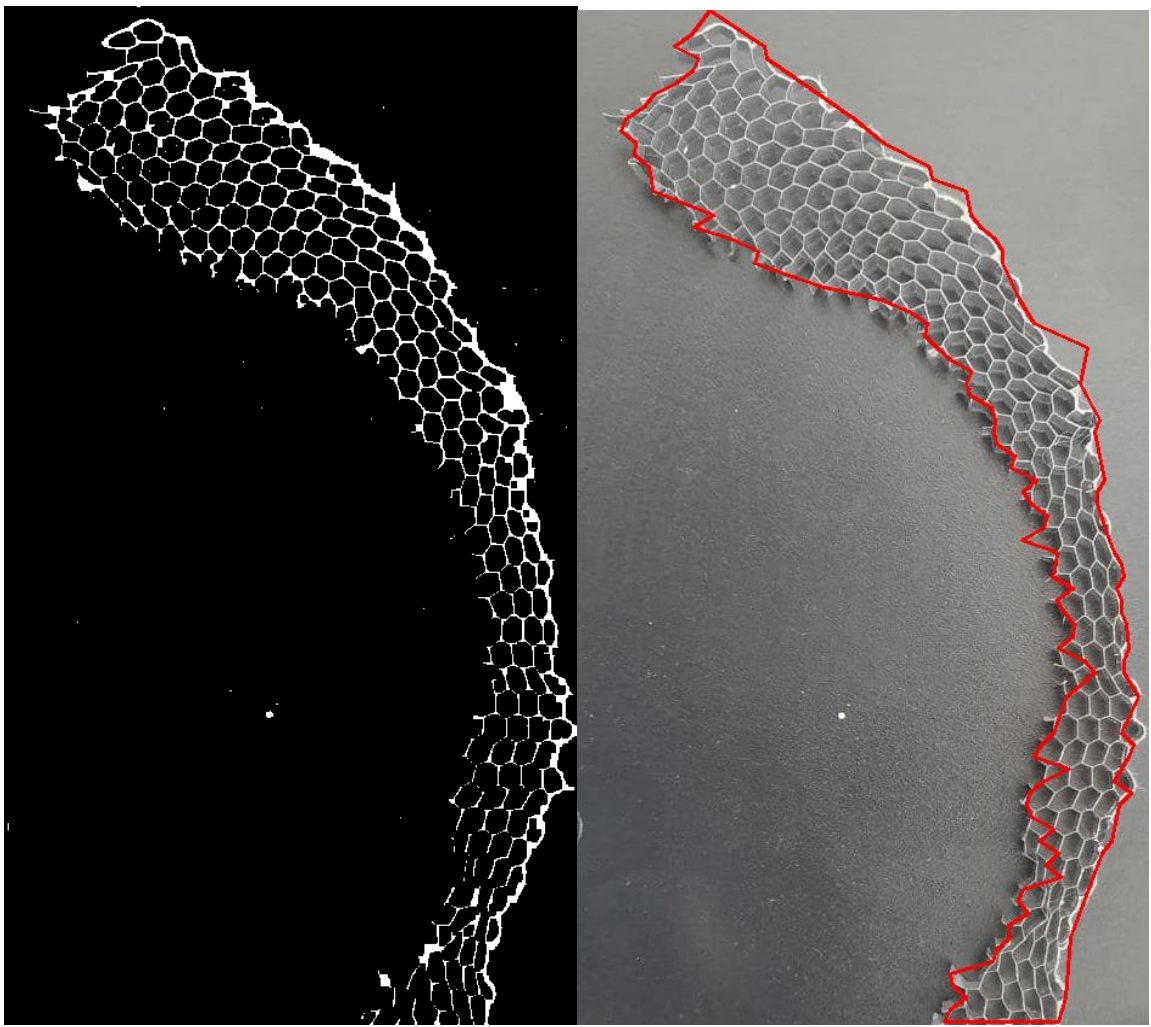

*Рис. 4. Бинаризованное изображение и обнаруженный контур.*

Как можно заметить на данном изображении присутствуют шумы в виде белых пикселей вне контура сотоблока, но алгоритм распознал их и не занес точки в контур.

После получения массива точек правого и левого края, можем построить контур данной заготовки. На рисунке 4 показан построенный контур данной заготовки.

Приведем еще несколько примеров обнаружения контура сотоблока (Рис. 5).

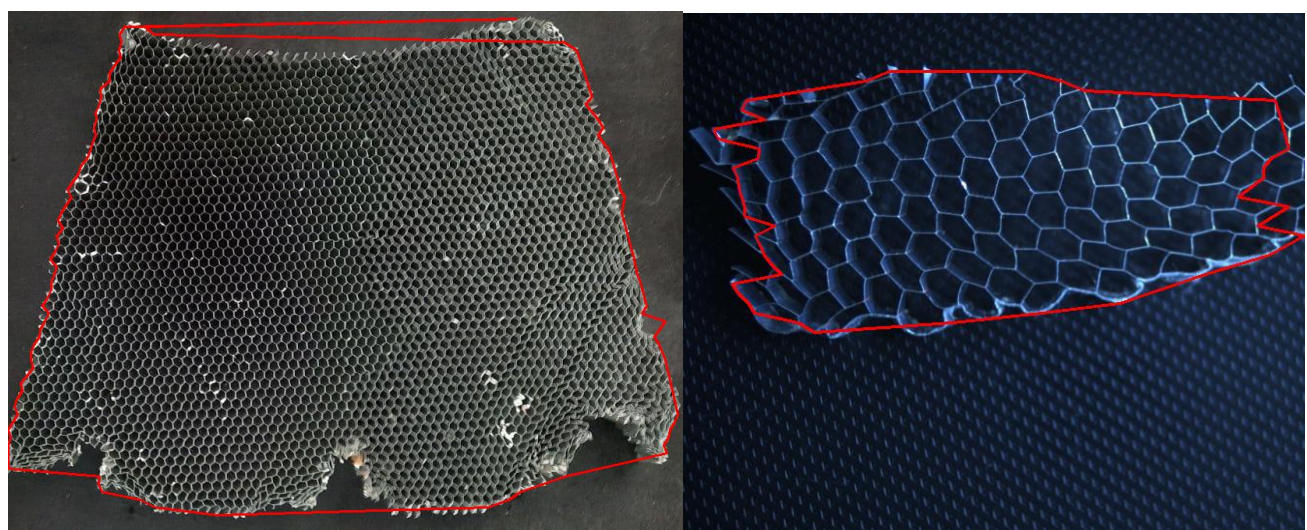

а)          б)

*Рис. 5. Контуры сотоблока.*

Как можно видеть данный алгоритм неплохо справляется с заготовками выпуклой формы, но с заготовкой вогнутой формы алгоритм работает не корректно или с большими неточностями (Рис. 5 а)). Продемонстрируем работу алгоритма на заготовке, представленной на рисунке 6.

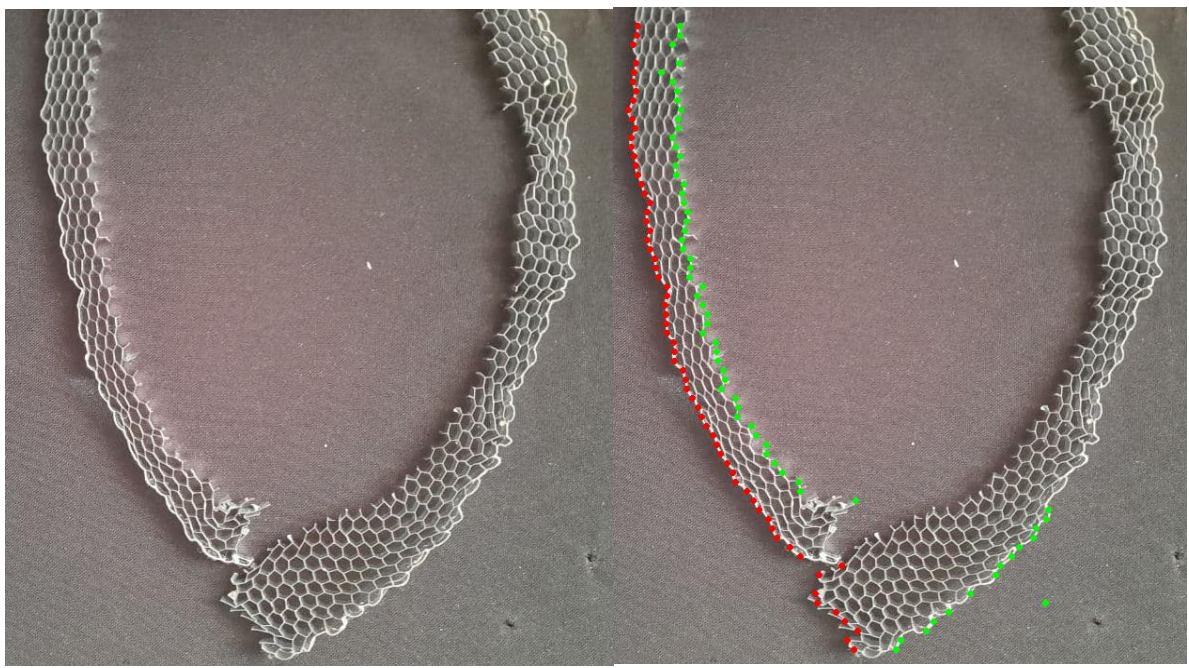

*Рис. 6. Заготовка выпуклой формы и найденные точки.*

Применим алгоритм к данному изображению, результат обнаруженных точек контура представлен на рисунке 6 справа.

Как можно видеть границы заготовки определены не верно, в результате мы получим контур, представленный на рисунке 7.

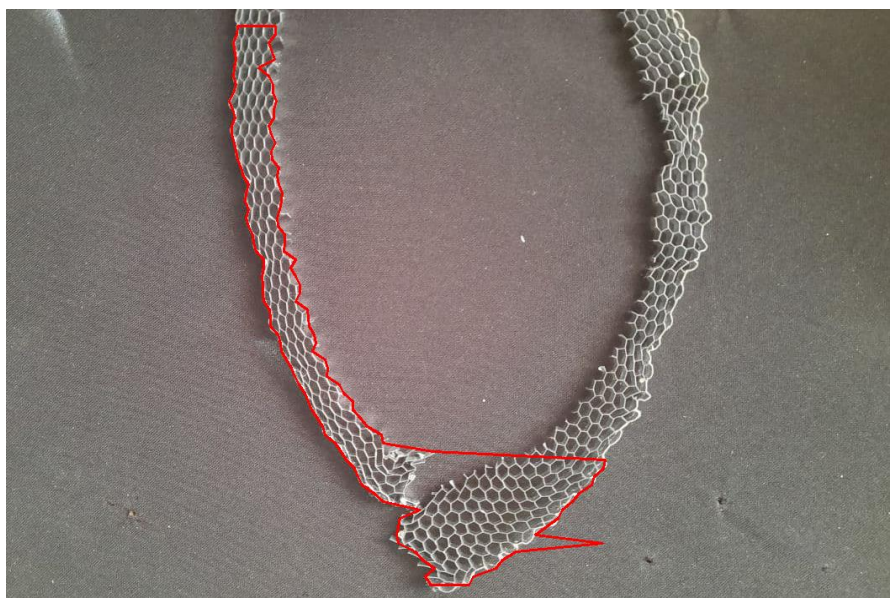

*Рис. 7. Контур заготовки сотоблока*

Контур определен верно только с левой стороны, правую часть алгоритм не распознал.

Данный алгоритм обладает высокой скоростью обработки изображения, достаточно точно определяет контуры сотоблока на выпуклой заготовке, в качестве параметра работы алгоритма выступает одна переменная, что упрощает работу с ним. Из недостатков стоит отметить некор-ректную работу с заготовками, которые имеют вогнутую форму, также не всегда корректно распознанные шумы.

## Алгоритм сканирования с помощью скользящей матрицы.

Данный алгоритм проверяет не каждый пиксель, как в предыдущем варианте, а анализирует пиксели, которые попали в скользящую матрицу, которая перемещается по изображению с заданным шагом.

Вычисляем среднее значение скользящей матрицы, если оно больше порогового значения, то приравниваем значение всех пикселей к 255, если меньше порогового значения, мы не производим никаких действий. На рисунке 8 показана схема работы данного алгоритма.

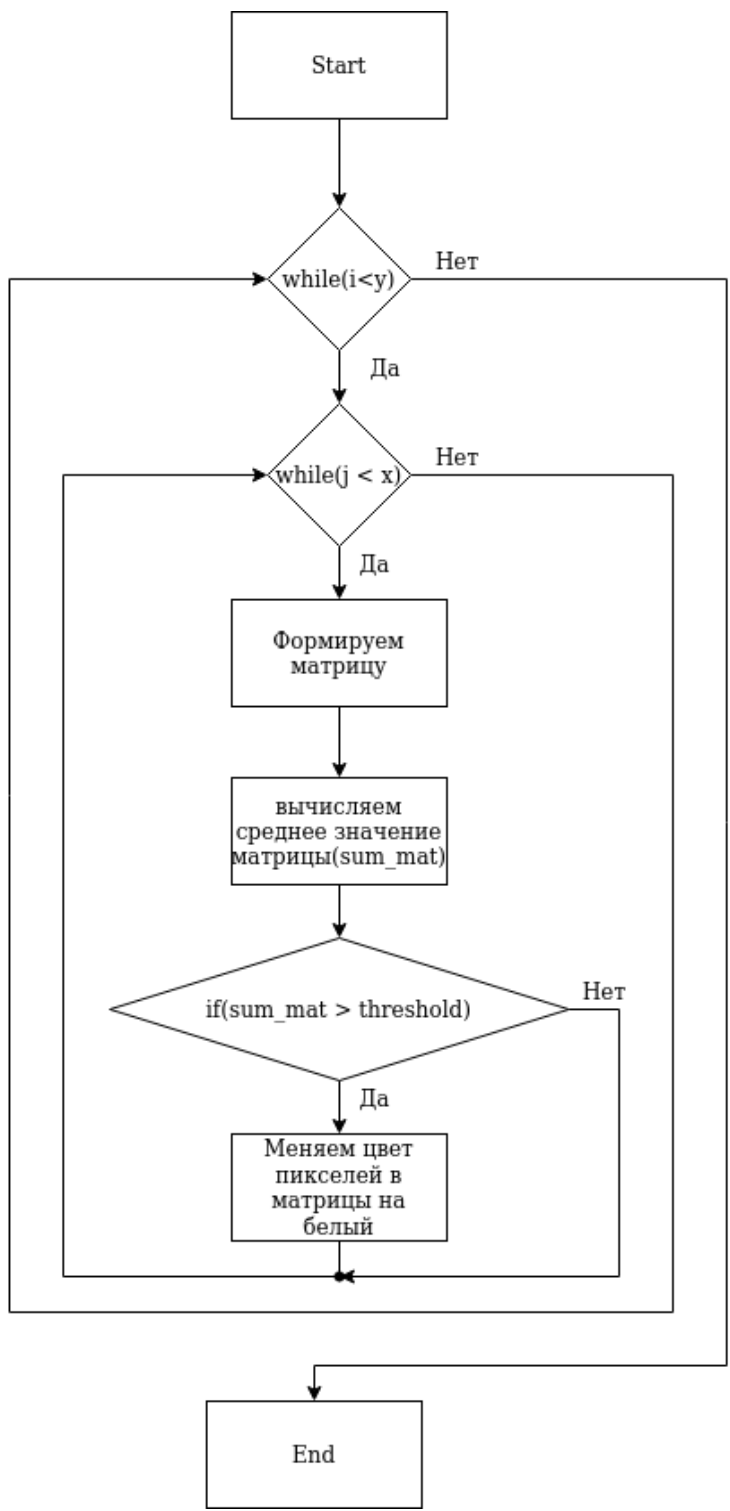

*Рис. 8. Блок схема алгоритма скользящей матрицы*

На рисунке 9 показаны исходное а), бинаризованное б) и изображение после обработки скользящей матрицы в).

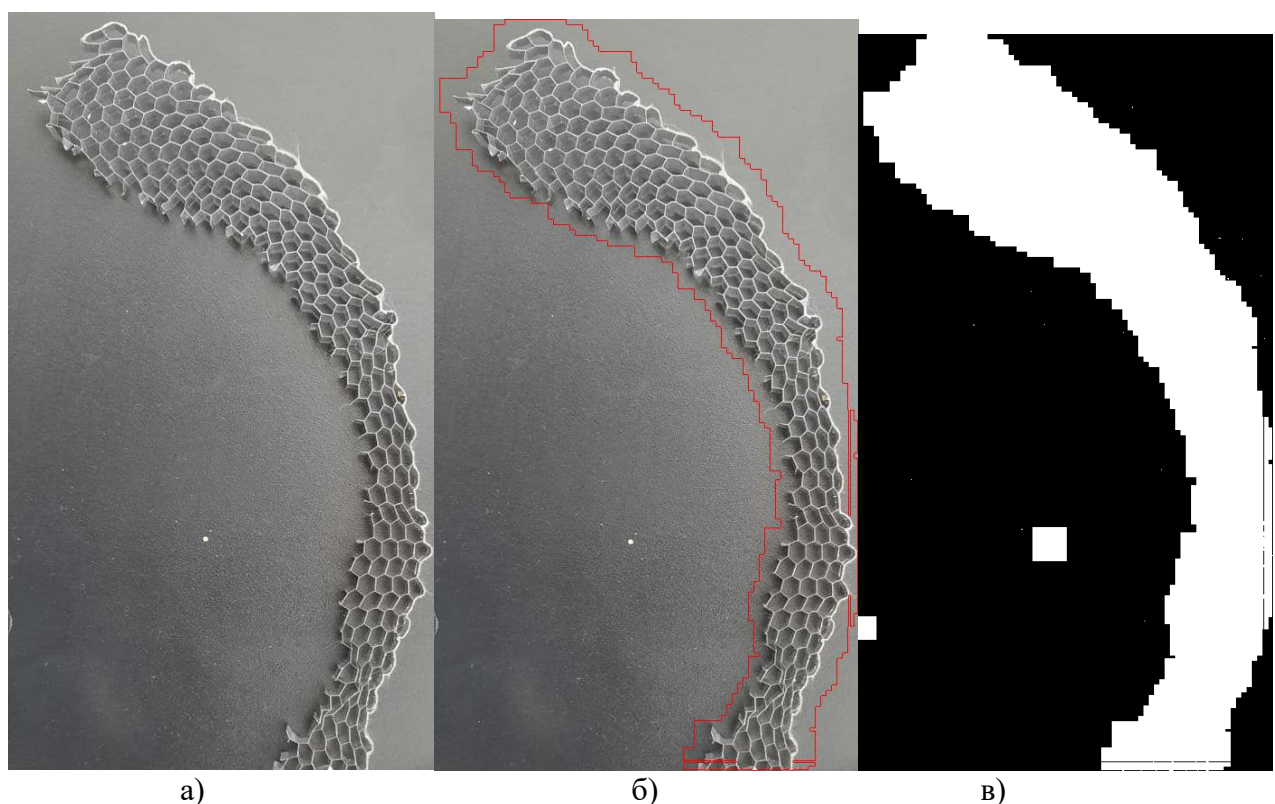

*Рис. 9. Алгоритм скользящей матрицы*

После того как была выполнена обработка скользящей матрицей, с помощью функции поиска контуров из библиотеки OpenCV, выполняем поиск контуров по внешним граням (Рис. 10). Далее выбираем из полученных контуров наибольший, тем самым избавляемся от тех контуров, которые описывают шум.

Данный алгоритм имеет несколько регулируемых параметров отвечающих за работу алгоритма:

1. core - параметр, отвечающий за размер матрицы
2. step - шаг скольжения матрицы (максимальный шаг рекомендуется выбирать не более 2*core)

Если уменьшить матрицу для примера выше, мы получим результат как на рисунке 11.

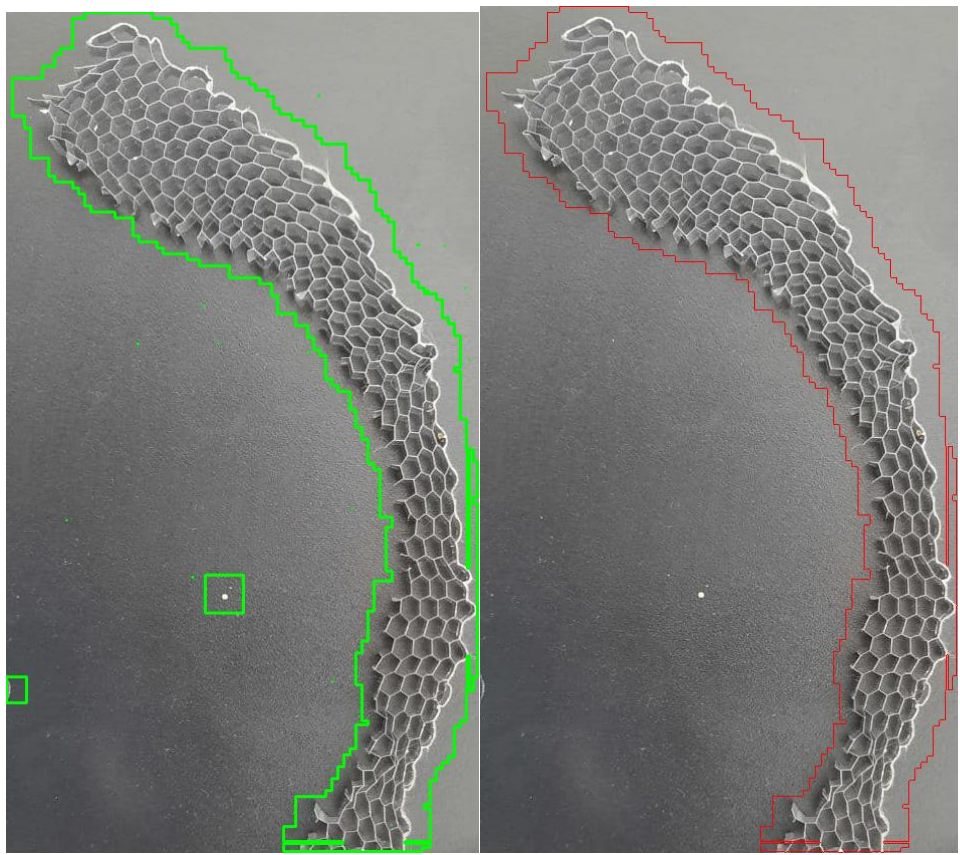

*Рис. 10. Обнаруженные контуры*

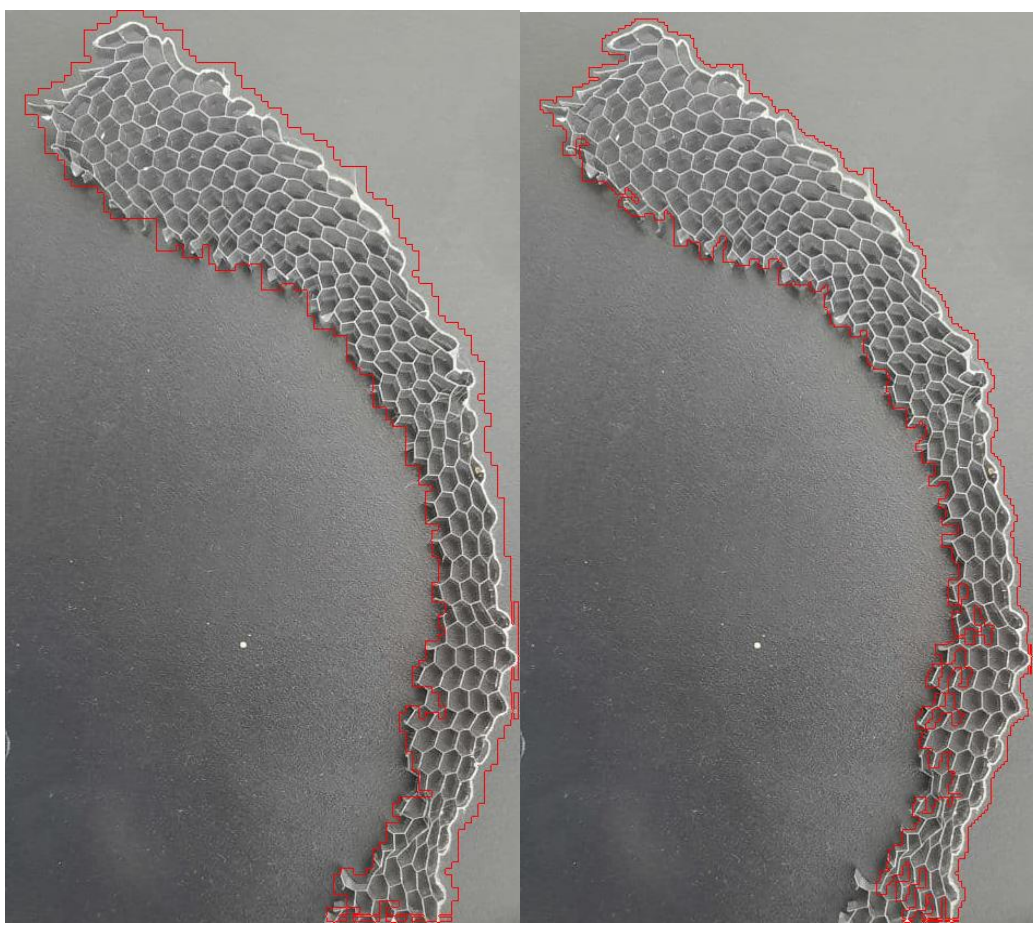

*Рис. 11. Контур заготовки, полученный матрицей с малым ядром.*

Как можно видеть контур приблизился к границам сотоблока, попробуем еще уменьшить ядро матрицы (Рис. 11 справа).

Как можно видеть контур прилегает к границе сотоблока, но в нижней части идет запутанный контур.

Если уменьшать шаг, то на выходе получим более плавный контур (Рис. 12)

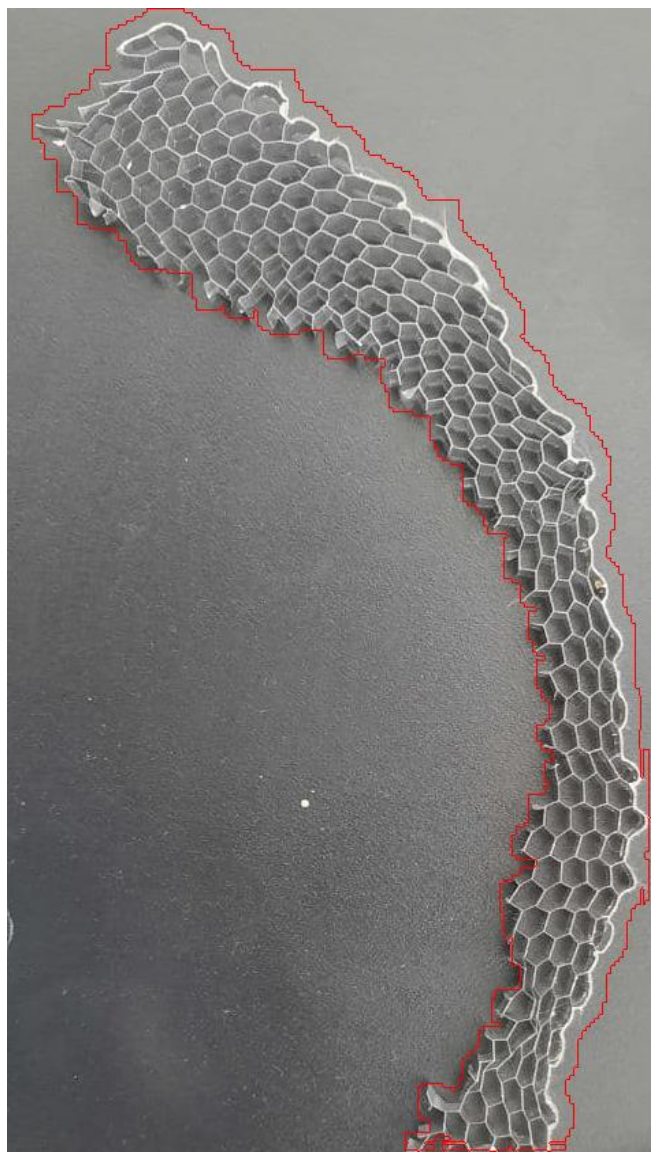

*Рис. 12. Контур сотоблока с малым шагом*

Также данный алгоритм способен обрабатывать заготовки сотоблоков вогнутой формы (Рис.13).

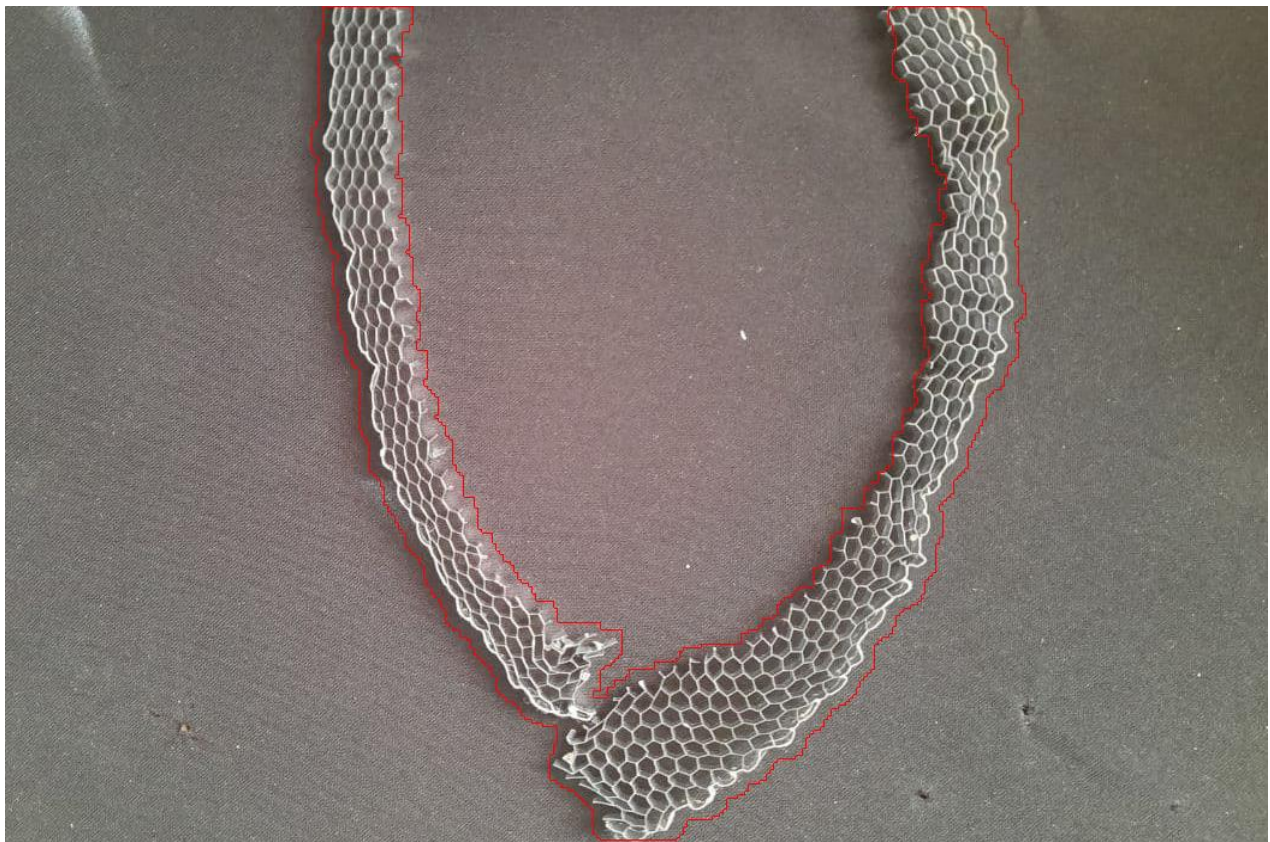

*Рис. 13. Контур сотоблока вогнутой формы*

**Заключение**

Данный алгоритм обладает рядом преимуществ перед первым алгоритмом
1. более гибкая настройка параметров
2. возможность обрабатывать заготовки любой формы
3. большая устойчивость к шуму

Из недостатков следует выделить увеличение времени на обработку изображения по сравнению с первым.

**СПИСОК ЛИТЕРАТУРЫ**